\documentclass[smallabstract,smallcaptions]{paper_c}

\usepackage{epsfig}
\usepackage{citesort}
\usepackage{amsmath}
\usepackage{amssymb}
\usepackage{color}
\usepackage{url}

\newlength{\figurewidth}
\newlength{\smallfigurewidth}

\usepackage{authblk}
\usepackage{layouts}

\usepackage{pgfplots}

\usepgfplotslibrary{external}
\begin{document}

\title
{\large
\textbf{Applicability limitations of differentiable full-reference image-quality metrics}
}

\title{Applicability limitations of differentiable full-reference image-quality metrics}
\author[1]{Siniukov Maksim}
\author[1,2]{Dmitriy Kulikov}
\author[1]{Dmitriy Vatolin}
\affil[1]{Lomonosov Moscow State University}
\affil[2]{Dubna State University}
\affil[ ]{\textit {\{maksim.siniukov,dkulikov,dmitriy\}@graphics.cs.msu.ru}}

\maketitle
\thispagestyle{empty}

\begin{abstract}
Subjective image-quality measurement plays a critical role in the development of image-processing applications. The purpose of a visual-quality metric is to approximate the results of subjective assessment. In this regard, more and more metrics are under development, but little research has considered their limitations. This paper addresses that deficiency: we show how image preprocessing before compression can artificially increase the quality scores provided by the popular metrics DISTS, LPIPS, HaarPSI, and VIF as well as how these scores are inconsistent with subjective-quality scores. We propose a series of neural-network preprocessing models that increase DISTS by up to 34.5\%, LPIPS by up to 36.8\%, VIF by up to 98.0\%, and HaarPSI by up to 22.6\% in the case of JPEG-compressed images. A subjective comparison of preprocessed images showed that for most of the metrics we examined, visual quality drops or stays unchanged, limiting the applicability of these metrics.
\end{abstract}

\Section{Introduction}
Image-quality measurement is critical to the development of image-processing algorithms such as compression, image enhancement, adaptive network streaming, and super-resolution. Despite the existence of objective quality metrics, visual quality is often a primary means of testing new algorithms. Perceptual quality is measurable through subjective evaluation, yielding the most-accurate results, but because subjective evaluation is complex and time consuming, it is unsuitable for frequent use; objective quality metrics are more practical. Yet to ensure precision, not just speed, metrics must correlate highly with perceptual results. \\
Full-reference (FR) metrics—metrics that require reference images to estimate the quality of images under test—fall into two categories: learning-based and non-learning-based (rules-based, heuristic, and algorithmic). Examples of non-learning-based metrics include NLPD \cite{NLPD}, HaarPSI \cite{HaarPSI}, and VIF \cite{VIF}. The developers of the normalized Laplacian-pyramid distance measure (NLPD) \cite{NLPD} adapted local luminance subtraction and local gain control by using a Laplacian pyramid to mimic an early visual system, yielding higher correlation with MOS than the well-known RMSE and MS-SSIM measures when evaluating compressed-image quality. The Haar wavelet-based perceptual-similarity index (HaarPSI) \cite{HaarPSI} uses the coefficients from a Haar wavelet decomposition to assess local similarities between two images to quantify the relative importance of image areas. Visual-information fidelity (VIF) \cite{VIF} estimates the information in a reference image and how much of it is extractable from the distorted image, thereby assessing a test image’s subjective quality.\\
Popular new learning-based metrics include PieAPP \cite{PieAPP}, LPIPS \cite{LPIPS}, and DISTS \cite{DISTS}. Perceptual image-error assessment through pairwise preference (PieAPP) \cite{PieAPP} is a learning-based method that predicts perceptual image error like human observers; it is founded on notice that a quality-estimation neural network is trainable on pairwise comparison data without explicit human-applied perceptual-error labels. Learned perceptual image-patch similarity (LPIPS) is a convolutional-neural-network-based estimator that uses features extracted by a pretrained VGG network. \cite{LPIPS} Deep image structure and texture similarity (DISTS) \cite{DISTS} is a novel CNN-based subjective quality metric that focuses on texture and structure similarity, as those characteristics greatly affect human perception. An independent research team \cite{antsiferova2022video} tested the aforementioned metrics and concluded that DISTS exhibits the highest correlation with subjective scores. The authors compared it to the methods mentioned above on the Live dataset and showed that DISTS has the highest correlation with MOS.\\ Correlation with subjective assessment often serves to assess a metric’s quality. The image-quality assessment benchmark (IQA) \cite{antsiferova2022video} showed that when tested only on compressed images (without preprocessing), DISTS achieved an SROCC correlation of 0.847 and LPIPS achieved 0.749. A different research team \cite{piq}, however, obtained different SROCC scores: 0.84 for PieAPP, 0.87 for HaarPSI, 0.81 for DISTS, and 0.67 for LPIPS. In addition to simple subjective comparison, more-in-depth studies of metric stability given various artificial distortions are necessary; certain distortions increase the metric values while decreasing or leaving unchanged the visual quality. In this paper we describe such distortions and analyze their effect on both subjective quality and metric scores. Today a growing number of perceptive quality metrics are under development, but in most cases, the researchers continue to use the old SSIM and PSNR. The reason for this tendency is a lack of knowledge regarding the limitations of state-of-the-art metrics, the focus of our paper.
\Section{Related work}
Hacking of VMAF, NIQE, and PSNR has undergone extensively study. Though tuning for objective metrics to achieve better image-comparison scores occurs throughout the industry, it can decrease the subjective quality, just as tuning for subjective metrics can decrease objective quality. The authors of \cite{preproc_proc} analyzed the effect of image preprocessing on visual quality and compared the results with objective metrics. In their experiments, subjective quality improved when the Gaussian-blurring PSNR decreased. Their results showed that a no-reference metric, NIQE, correlated better with subjective scores for preprocessed images. But NIQE can also be inconsistent. In other research \cite{Kulikov2019}, the authors demonstrated that NIQE has difficulty estimating the quality of dark and highly textured frames. \\
Many image-processing algorithms implement tuning filters, which are especially popular in image encoders. For example, x264 and x265 provide tuning options for PSNR and SSIM, and recently, the developers of libaom implemented VMAF tuning \cite{VMAF_rd}. Their paper showed that preprocessing enabled a substantial compression gain (the bitrate fell by 37.91\% for the same VMAF), while rate-distortion optimization enabled a much smaller gain (4.69\%). \\The authors of \cite{hackingvmaf} observed that histogram equalization and unsharp masking can artificially increase VMAF by 5–6\%; other researchers also proved VMAF’s vulnerability to preprocessing \cite{ozer_2020}. Elsewhere, \cite{HackingVMAFneg} showed that VMAF Neg values are unalterable using this approach, but both VMAF and VMAF Neg proved vulnerable when using a different pipeline in which images underwent compression with H.264. The authors of \cite{ProxIQA} used a generative-adversarial approach to improve the VMAF scores for a learnable image-compression algorithm. \\
Many studies, including \cite{hackingNR_Ekaterina} and \cite{NR_notour}, have searched for no-reference-metric vulnerabilities. In \cite{kettunen2019lpips}, the authors showed that an iterative algorithm can decrease LPIPS scores in a special way that yields negatively correlated MOS and LPIPS scores. Several methods for finding C  for PSNR metrics have also emerged, including \cite{psnr_lp}. Another study generated images with equal SSIM values (as a proxy for visual quality) but different PSNR values \cite{psnr_eq_ssim}. Because the metrics we discuss here are newer, they are less well researched. \cite{psnr_ssim_bovik}

\begin{center}
    \begin{figure}[h]
     \centering
     \includegraphics[width = 1.\linewidth]{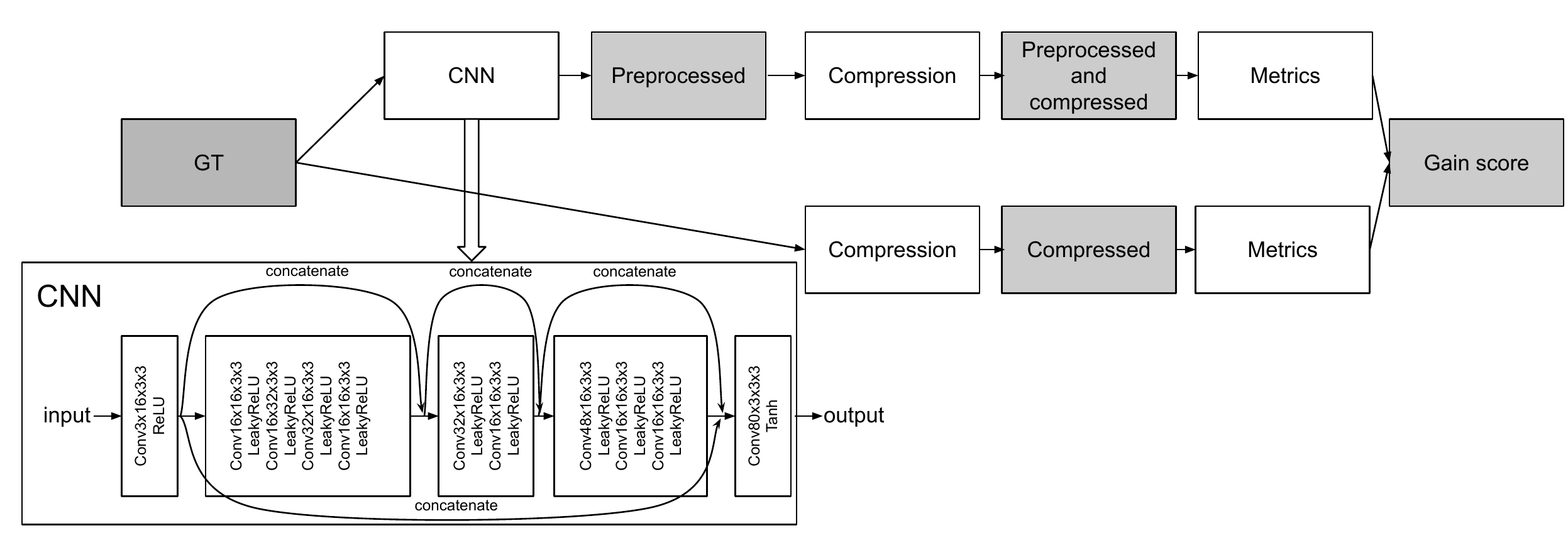}
     \caption{\label{fig:pipeline}%
    Hacking pipeline.}
    \end{figure}
\end{center}

\begin{center}
    \begin{figure}[h]
     \centering
      \scalebox{0.35}{\input{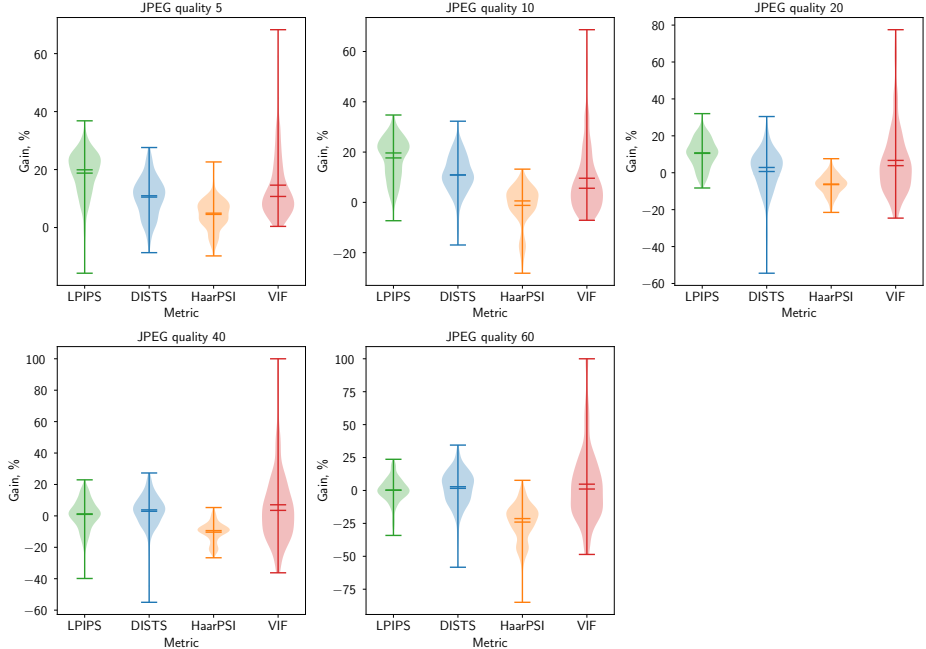}}
    \caption{\label{fig:gain}%
    Metric-gain distribution.}
    \end{figure}
\end{center}

\begin{center}
    \begin{figure}[h]
     \centering
     \includegraphics[width = 1.\linewidth]{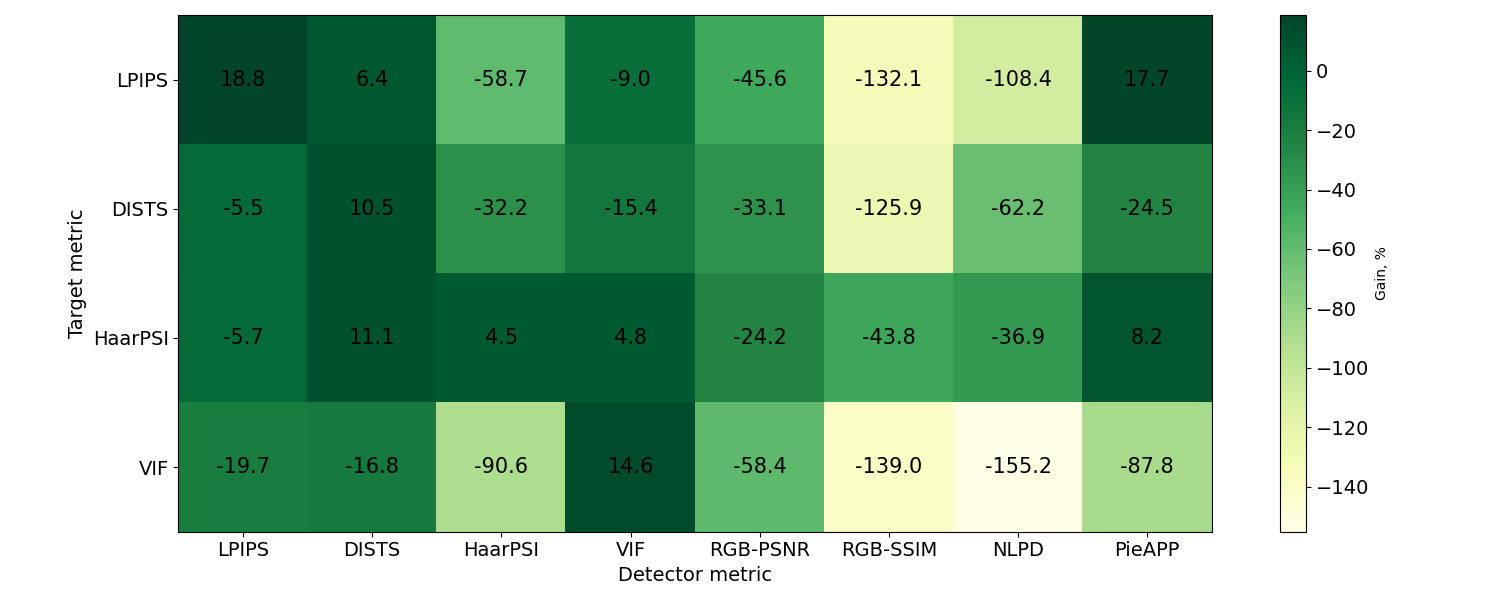}
\caption{\label{fig:heatmap}%
    Hacking-detection heatmap.}
    \end{figure}
\end{center}

%\begin{center}
%    \begin{figure}[h]
%     \centering
%     %\includegraphics[width = 1.\linewidth]{Figures/dataset_size_lineplot_jpeg.png}
%     \scalebox{0.9}{\input{Figures/dataset_size_lineplot_jpeg1.pgf}}
%    \caption{\label{fig:datasetsz}%
%    Test-loss dependence on training-dataset size.}
%\end{figure}
%\end{center}

\begin{table}[tp]
\begin{center}
\caption{\label{tab:datasetsz}%
Test-loss dependence on training-dataset size.}
{
\renewcommand{\baselinestretch}{1}\footnotesize
\begin{tabular}{|c|c|c|c|c|c|c|c|c|c|}
\hline
%\multicolumn{1}{|c|}{} &
dataset size & 100& 250& 500& 1000& 200& 5000& 11000& 20000& 30000 \\
\hline
DISTS& 0.5496& 0.503& 0.456& 0.441& 0.429& 0.427& 0.423& 0.4225& 0.4222 \\
\hline
\end{tabular}}
\end{center}
\end{table}

\Section{Proposed method}
Our goal was to find vulnerabilities that make metric inapplicable to compressed images, as most real-world images are compressed (using JPEG). For each metric we a found universal image transformation such that given any input ground-truth (GT) image, we can generate a pair of images: one JPEG compressed (C) and the other transformed using a one-for-all-images neural network and then compressed using JPEG (TC). If the target metric ranks TC higher than C while subjective scores and other metrics do the opposite, the target metric is unsuitable for subjective quality assessment when used with JPEG compression. But codec developers can include this preprocessing step in their methods to boost the outputs of vulnerable metrics and place higher in benchmarks. Since JPEG is the most common image-compression algorithm, it is of utmost importance for metric comparison. \\
We propose a preprocessing algorithm based on a ResNet-like lightweight CNN architecture, shown in Figure \ref{fig:pipeline}. The total number of weights is 32,707 and the model size is 133 KB (136,681 bytes), yielding high performance (up to 10.95 fps on Nvidia RTX 2080Ti GPU; 0.17fps on Intel Core i7-3770 CPU; 0.07fps on Snapdragon 855 mobile CPU for FullHD images) and making this approach amenable to online use. Other compression methods can thus include this preprocessing as well. \\
The second step of our method is compression. Because gradient propagation is necessary for training the preprocessor, we chose differentiable compression algorithms. Also, since regular JPEG implementations are undifferentiable, we used the differentiable DiffJPEG \cite{shin2017diffjpeg}, which mimics typical JPEG behavior. We chose the implementation from \cite{diffjpeg_github} for our training pipeline. Our last step is FR-metric calculation. We employed each metric on a TC image for evaluation and a GT image without preprocessing for reference. We targeted our loss function to decrease the difference in scores between two given images (a lower metric value corresponds to higher subjective quality). Images generated in this way exhibited abnormally high metric values that exceed those of the original compressed images. This result raised the question of whether the evaluation correlates with subjective scores and other metric scores. \\
All preprocessing models underwent training on 11,000 images from Vimeo90K \cite{vimeo90k} and were tested on 1,500 images from different Vimeo90K sequences; this dataset is widely used for image processing and has served in numerous research works \cite{vimeo90k_cite1}, \cite{vimeo90k_cite2}. We chose the training-dataset size of 11,000 after considering how different sizes affect metric gains on the test set. Table~\ref{tab:datasetsz} presents the results as a line plot, showing that metrics plateau at a dataset size of 11,000. We trained the neural networks using the early-stopping technique: the maximum epoch count was 50, but training stopped if the test loss failed to improve over five successive epochs. %The code is at LINK IS HIDDEN FOR BLIND REVIEWhttps://github.com/havent-invented/MetricsHacking/.
%\begin{table}[tp]
%\begin{center}
%\caption{\label{tab:fps}%
%Preprocessing speed on a single FullHD image}
%{
%\renewcommand{\baselinestretch}{1}\footnotesize
%\begin{tabular}{|c|c|c|c|}
%\hline
%\multicolumn{1}{|c|}{} &
%GPU & CPU & mobile\\
%\hline
%ms/frame &91.3 &5940 &14400\\
%fps &10.95 &0.17 &0.07 \\
%\hline
%\end{tabular}}
%\end{center}
%\end{table}

\Section{Practical study}
We trained our preprocessing neural networks with DiffJPEG, but compression usually employs the nondifferentiable JPEG format. Therefore, we tested the transferability of the experimental results to various nondifferentiable JPEG implementations—in particular, OpenCV \cite{opencv} and PIL \cite{PILpython}. The test pipeline began by training the preprocessing models with DiffJPEG under different quality parameters. It then calculated results for our target metrics using the trained preprocessing model for an image compressed with standard JPEG and, as a reference, for a GT image without preprocessing. Therefore, we ensured that the metrics yield abnormally high scores when images are compressed with standard JPEG, which is a restriction of subjective IQA metrics. Table~\ref{transfer} presents our results; metric gains are lower than for metrics based on DiffJPEG. Therefore, models trained using DiffJPEG can serve in practice with real JPEG compression, and they can also serve in specially designed encoders as a preprocessing step that yields higher metric scores when comparing JPEG encoders. \\
In addition to image compression, this approach works for video compression, too, as frames are compressed using image-compression algorithms; including the preprocessing step in those algorithms may yield unreasonably high scores in codec benchmarks. The AV1 codec has already implemented this preprocessing step through the “-tune vmaf” option \cite{VMAF_rd} to boost scores from the vulnerable full-reference VMAF metric. AV1’s VMAF tuning, however, is based on an iterative algorithm that performs an exhaustive search, slowing compression dramatically, whereas ours is the first one-pass approach and has almost no effect on compression speed.

\begin{table}[tp]
\begin{center}
\caption{\label{transfer}%
Transferability on real JPEG of quality 10 }
{
\renewcommand{\baselinestretch}{1}\footnotesize
\begin{tabular}{|c|c|c|c|c|}
\hline
\multicolumn{1}{|c|}{} &
DISTS & LPIPS & HaarPSI & VIF\\
\hline
average gain&6.31\% & 18.94\%& 2.03\% &20.40\%\\
\hline
\end{tabular}}
\end{center}
\end{table}

\begin{center}
    \begin{figure}[h]
     \centering
      \includegraphics[width = 1.\linewidth]{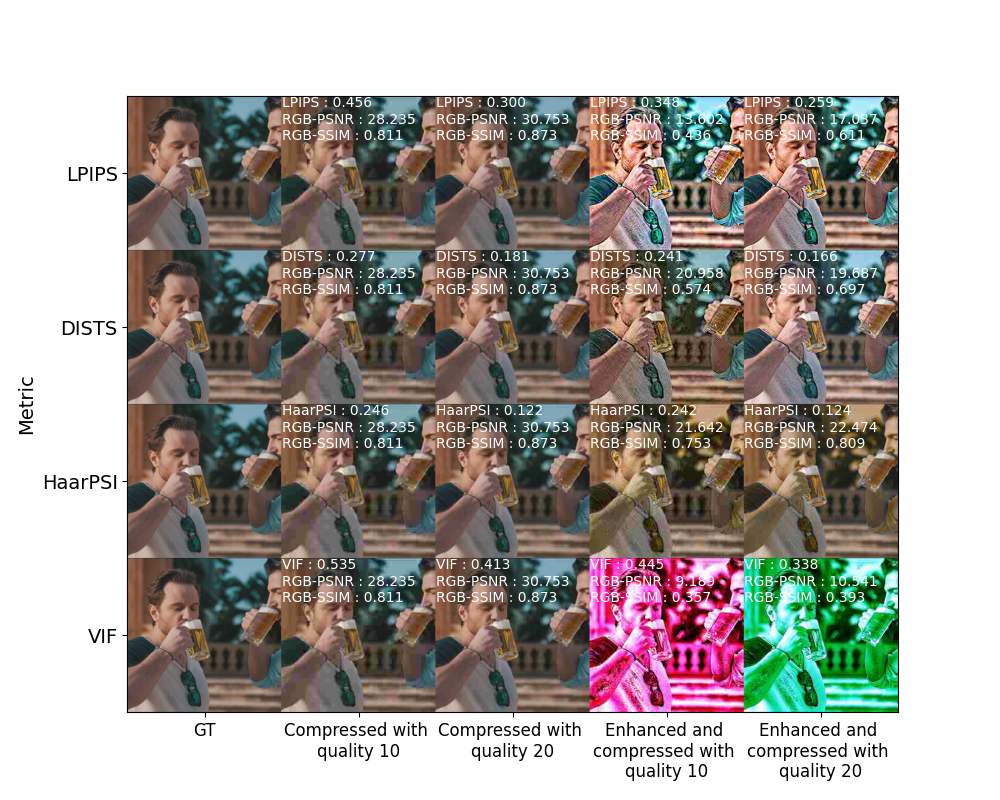}
    \caption{\label{fig:patches}%
    Metric-hacking examples.}
    \end{figure}
\end{center}

\Section{Subjective evaluation}
IQA metrics are designed to imitate the subjective judgment of the human eye, so proving that their scores correlate negatively with subjective scores would indicate they have a major downside. To verify that metrics evaluate our generated image pairs differently relative to subjective human-perception scores, we chose several image pairs and asked volunteers to decide which image in each pair is better. 531 people participated in the subjective evaluation. To ensure they understood the assignment and did their job correctly, we included another 2 images for which only one option was reasonable.\\ We used Bradley-Terry, a common model for converting the results of a pairwise comparison into a single MOS measure; Table~\ref{tab:subj} shows the resulting Bradley-Terry scores for our experiment. 
They are much lower than when we tested the metrics on images compressed without any preprocessing. Participants preferred images that were not preprocessed, resulting in low correlation between MOS scores and the values for the metrics we examined. Therefore, these metrics are unsuitable for IQA subjective evaluation of compressed images owing to their anticorrelation with human assessment.\\
The worst metrics in terms of matching MOS proved to be VIF, while the best was DISTS. 
%Table~\ref{tab:subj} compares our test results and unhacked-metric performance on the benchmarks of independent researchers. 
The hacked metrics ceased to correspond to human perception as their scores increased, because the subjective scores dropped. Examples of preprocessed images and their compressed versions appear in Figure \ref{fig:patches}
%['DISTS', 'LPIPS', 'HaarPSI', 'VIFLoss']
%[0.05213179519524486, 0.12669003805315407, -0.01035415491428474, 0.12539977643557623]
%5.21317952, 12.66900381, -1.03541549, 12.53997764
%[59, 72, 46, 61]
%[80, 80, 80, 80]

\begin{table}[tp]
\begin{center}
\caption{\label{tab:subj}%
Subjective evaluation results}
{
\renewcommand{\baselinestretch}{1}\footnotesize
\begin{tabular}{|c|c|c|c|c|}
\cline{2-5}
\multicolumn{1}{c|}{~}&
\multicolumn{1}{c|}{metric} &
\multicolumn{1}{c|}{subjective} &
\multicolumn{2}{c|}{hacked images percentage chosen as the best}\\
\cline{4-5}
\multicolumn{1}{c|}{} & %
gain & Bradley-Terry(TC/C) & by metric & by people\\
\hline
DISTS &5.21\% &1.176/0.583 & 73.8\% &22.4\%\\
LPIPS &12.67\% &0.714/0.357 & 90.0\% &32.2\%\\
HaarPSI &-1.03\% &1.367/0.683 & 57.5\% &17.9\%\\
VIF &12.54\% &3.857/1.929 & 76.3\% &1.3\%\\
\hline
\end{tabular}}
\end{center}
\end{table}

\Section{Hacking generalization}
We found that a gain in metric values emerges for 93.6\% of our 1,500 test images compressed using DiffJPEG at quality level 5 for the case of DISTS, 97.8\% for LPIPS, 83\% for HaarPSI, and 100\% for VIF. Some images, however, saw a lower gain than others, and some saw a much higher gain. The supplementary materials illustrate both kinds of outliers. DISTS and LPIPS showed no increased for highly contrasting images and increased the most when the images were noncontrasting. HaarPSI exhibited no increase when the images were extremely bright. Therefore, our approach appears  universal, as it worked on nearly all the images we tested. Figure \ref{fig:gain} presents the gain distribution across the dataset as a violin plot, revealing that DISTS has a low standard deviation of gains—meaning it resists hacking—whereas LPIPS is susceptible to hacking.%DISTS0.928 LPIPS0.986 HaarPSI0.528 VIF0.73 for quality 10
\Section{Hacking detection}
As mentioned above, developers may include hacking algorithms in their algorithms, so the task of detecting such algorithms becomes relevant. In practice, hacking detection can be implemented as follows: if the results of a subjective evaluation anticorrelate with metric results, as we described in the previous section, the metric was hacked. But subjective evaluations are time consuming and require numerous participants. Therefore, they are impractical for routine hacking detection. Other metric values, however, may reveal whether the target metric was hacked. We discovered how hacking one metric affects the output of other metrics, shown as a heat map in Figure \ref{fig:heatmap}. Generally, other metrics tend to indicate lower-quality images by providing lower scores, but, for example, DISTS is immune to LPIPS hacking, possibly because they have similar structures. PSNR and SSIM are the best hacking detectors.

\Section{Future work}
Another metric vulnerability of interest is low-visibility hacking that only slightly changes the image, leaving no artifacts and leaving other metric values unchanged. In such cases, hacking could go unnoticed. The first step to exploit this vulnerability may be development of a preprocessing model that increases all IQA metrics at once, leaving no evidence of hacking. In this case, subjective scores remain unchanged, whereas target metrics yield higher scores for the images.

\Section{Conclusion}
The DISTS, LPIPS, HaarPSI, and VIF full-reference image-quality metrics are not versatile, as certain universal transformations can increase their output values and decrease their correlation with MOS. Thus, they can provide inaccurate image quality assessment and should not serve as the basis for decision making in quality evaluation%Thus, they can provide inaccurate image-quality assessment and should not serve as the main standard.
We showed that hacking can increase these metrics by up to 98.0\%; the lowest gain was in the case of HaarPSI and reached of original values. Their values may be misleading when evaluating compression algorithms, so subjective assessment is still necessary. The traditional PSNR and SSIM, however, are immune to such increases, making them a good signal for hacking detection, but their correlation with MOS is too low to make them useful as subjective-quality measures. Therefore, development of novel universal visual-quality-assessment methods remains a relevant problem. \cite{wang2008maximum}. The developers of VMAF, for instance, have proposed a hacking-resistant version of their metric, VMAF Neg. Similarly, less vulnerable versions of the metrics we explored in this paper may likewise emerge.
\Section{References}
\bibliographystyle{IEEEbib}
\bibliography{main}

\begin{thebibliography}{10}

\bibitem{NLPD}
Valero Laparra, Johannes Ballé, Alexander Berardino, and Eero Simoncelli,
\newblock ``Perceptual image quality assessment using a normalized laplacian
  pyramid,''
\newblock {\em Electronic Imaging}, vol. 2016, pp. 1--6, 02 2016.

\bibitem{HaarPSI}
Rafael Reisenhofer, Sebastian Bosse, Gitta Kutyniok, and Thomas Wiegand,
\newblock ``A haar wavelet-based perceptual similarity index for image quality
  assessment,''
\newblock {\em Signal Processing: Image Communication}, vol. 61, pp. 33--43,
  2018.

\bibitem{VIF}
H.R. Sheikh and A.C. Bovik,
\newblock ``Image information and visual quality,''
\newblock {\em IEEE Transactions on Image Processing}, vol. 15, no. 2, pp.
  430--444, 2006.

\bibitem{PieAPP}
Ekta Prashnani, Hong Cai, Yasamin Mostofi, and Pradeep Sen,
\newblock ``Pieapp: Perceptual image-error assessment through pairwise
  preference,''
\newblock in {\em 2018 IEEE/CVF Conference on Computer Vision and Pattern
  Recognition}, 2018, pp. 1808--1817.

\bibitem{LPIPS}
Richard Zhang, Phillip Isola, Alexei~A. Efros, Eli Shechtman, and Oliver Wang,
\newblock ``The unreasonable effectiveness of deep features as a perceptual
  metric,''
\newblock {\em CoRR}, vol. abs/1801.03924, 2018.

\bibitem{DISTS}
Keyan Ding, Kede Ma, Shiqi Wang, and Eero~P. Simoncelli,
\newblock ``Image quality assessment: Unifying structure and texture
  similarity,''
\newblock {\em IEEE Transactions on Pattern Analysis and Machine Intelligence},
  vol. 44, no. 5, pp. 2567--2581, 2022.

\bibitem{antsiferova2022video}
Anastasia Antsiferova, Sergey Lavrushkin, Maksim Smirnov, Aleksandr Gushchin,
  Dmitriy Vatolin, and Dmitriy Kulikov,
\newblock ``Video compression dataset and benchmark of learning-based
  video-quality metrics,''
\newblock in {\em Thirty-sixth Conference on Neural Information Processing
  Systems Datasets and Benchmarks Track}, 2022.

\bibitem{piq}
PIQ,
\newblock ``Pytorch image quality),''
  \url{https://github.com/photosynthesis-team/piq}.

\bibitem{preproc_proc}
Jayachandra Chilukamari, Sampath Kannangara, and Grant Maxwell,
\newblock ``Investigation of the effectiveness of video quality metrics in
  video pre-processing,''
\newblock in {\em 2013 IEEE Third International Conference on Consumer
  Electronics ¿ Berlin (ICCE-Berlin)}, 2013, pp. 1--5.

\bibitem{Kulikov2019}
A.~Zvezdakova, D.~Kulikov, D.~Kondranin, and D.~Vatolin,
\newblock ``Barriers towards no-reference metrics application to compressed
  video quality analysis: On the example of no-reference metric niqe,''
\newblock {\em CEUR Workshop Proceedings}, vol. 2485, pp. 22--27, 2019.

\bibitem{VMAF_rd}
Sai Deng, Jingning Han, and Yaowu Xu,
\newblock ``Vmaf based rate-distortion optimization for video coding,''
\newblock in {\em 2020 IEEE 22nd International Workshop on Multimedia Signal
  Processing (MMSP)}, 2020, pp. 1--6.

\bibitem{hackingvmaf}
Anastasia Antsiferova, Dmitriy~L. Kulikov, Dmitriy~S. Vatolin, and Sergey
  Zvezdakov,
\newblock ``Video distortion method for {VMAF} quality values increasing,''
\newblock {\em CoRR}, vol. abs/1907.04807, 2019.

\bibitem{ozer_2020}
Jan Ozer,
\newblock ``Vmaf is hackable: What now?,'' Apr 2020.

\bibitem{HackingVMAFneg}
Maksim Siniukov, Anastasia Antsiferova, Dmitriy Kulikov, and Dmitriy Vatolin,
\newblock ``Hacking vmaf and vmaf neg: Vulnerability to different preprocessing
  methods,''
\newblock in {\em 2021 4th Artificial Intelligence and Cloud Computing
  Conference}, New York, NY, USA, 2022, AICCC '21, p. 89–96, Association for
  Computing Machinery.

\bibitem{ProxIQA}
Li-Heng Chen, Christos~G. Bampis, Zhi Li, Andrey Norkin, and Alan~C. Bovik,
\newblock ``Proxiqa: A proxy approach to perceptual optimization of learned
  image compression,''
\newblock {\em IEEE Transactions on Image Processing}, vol. 30, pp. 360--373,
  2021.

\bibitem{hackingNR_Ekaterina}
Ekaterina Shumitskaya, Anastasia Antsiferova, and Dmitriy Vatolin,
\newblock ``Universal perturbation attack on differentiable no-reference image-
  and video-quality metrics,''
\newblock {\em The British Machine Vision Conference}, 2022.

\bibitem{NR_notour}
Weixia Zhang, Dingquan Li, Xiongkuo Min, Guangtao Zhai, Guodong Guo, Xiaokang
  Yang, and Kede Ma,
\newblock ``Perceptual attacks of no-reference image quality models with
  human-in-the-loop,'' 2022.

\bibitem{kettunen2019lpips}
Markus Kettunen, Erik H{\"a}rk{\"o}nen, and Jaakko Lehtinen,
\newblock ``E-lpips: robust perceptual image similarity via random
  transformation ensembles,''
\newblock {\em arXiv preprint arXiv:1906.03973}, 2019.

\bibitem{psnr_lp}
Mahmood Sharif, Lujo Bauer, and Michael Reiter,
\newblock ``On the suitability of lp-norms for creating and preventing
  adversarial examples,''
\newblock 06 2018, pp. 1686--16868.

\bibitem{psnr_eq_ssim}
Frank~M. Ciaramello and Amy~R. Reibman,
\newblock ``{Supplemental subjective testing to evaluate the performance of
  image and video quality estimators},''
\newblock in {\em Human Vision and Electronic Imaging XVI}, Bernice~E. Rogowitz
  and Thrasyvoulos~N. Pappas, Eds. International Society for Optics and
  Photonics, 2011, vol. 7865, p. 78650Q, SPIE.

\bibitem{psnr_ssim_bovik}
Zhou Wang, A.C. Bovik, H.R. Sheikh, and E.P. Simoncelli,
\newblock ``Image quality assessment: from error visibility to structural
  similarity,''
\newblock {\em IEEE Transactions on Image Processing}, vol. 13, no. 4, pp.
  600--612, 2004.

\bibitem{shin2017diffjpeg}
Richard Shin and Dawn Song,
\newblock ``Jpeg-resistant adversarial images,''
\newblock in {\em NIPS 2017 Workshop on Machine Learning and Computer
  Security}, 2017, vol.~1, p.~8.

\bibitem{diffjpeg_github}
Mlomnitz,
\newblock ``Diffjpeg,'' \url{https://github.com/mlomnitz/DiffJPEG}.

\bibitem{vimeo90k}
Tianfan Xue, Baian Chen, Jiajun Wu, Donglai Wei, and William~T. Freeman,
\newblock ``Video enhancement with task-oriented flow,''
\newblock {\em International Journal of Computer Vision}, vol. 127, no. 8, pp.
  1106--1125, Aug 2019.

\bibitem{vimeo90k_cite1}
Xintao Wang, Kelvin~C.K. Chan, Ke~Yu, Chao Dong, and Chen Change~Loy,
\newblock ``Edvr: Video restoration with enhanced deformable convolutional
  networks,''
\newblock in {\em Proceedings of the IEEE/CVF Conference on Computer Vision and
  Pattern Recognition (CVPR) Workshops}, June 2019.

\bibitem{vimeo90k_cite2}
Wenbo Bao, Wei-Sheng Lai, Chao Ma, Xiaoyun Zhang, Zhiyong Gao, and Ming-Hsuan
  Yang,
\newblock ``Depth-aware video frame interpolation,''
\newblock in {\em Proceedings of the IEEE/CVF Conference on Computer Vision and
  Pattern Recognition (CVPR)}, June 2019.

\bibitem{opencv}
Opencv,
\newblock ``Opencv/opencv: Open source computer vision library,''
  \url{https://github.com/opencv/opencv}.

\bibitem{PILpython}
Python-Pillow,
\newblock ``Python-pillow/pillow: The friendly pil fork (python imaging
  library),'' \url{https://github.com/python-pillow/Pillow}.

\bibitem{wang2008maximum}
Zhou Wang and Eero~P Simoncelli,
\newblock ``Maximum differentiation (mad) competition: A methodology for
  comparing computational models of perceptual quantities,''
\newblock {\em Journal of Vision}, vol. 8, no. 12, pp. 8--8, 2008.

\end{thebibliography}

\end{document}